\documentclass[10pt,twocolumn,letterpaper]{article}

\usepackage{iccv}
\usepackage{times}
\usepackage{epsfig}
\usepackage{graphicx}
\usepackage{amsmath}
\usepackage{amssymb}

\usepackage[pagebackref=true,breaklinks=true,letterpaper=true,colorlinks,bookmarks=false]{hyperref}

\iccvfinalcopy 

\ificcvfinal\fi

\begin{document}

\title{SALA: Soft Assignment Local Aggregation for Parameter Efficient \\3D Semantic Segmentation}

\author{Hani Itani \qquad Silvio Giancola \qquad  Ali Thabet \qquad Bernard Ghanem\\
King Abdullah University of Science and Technology (KAUST)\\
{\tt\small \{hani.itani.2, silvio.giancola, ali.thabet, bernard.ghanem\}@kaust.edu.sa}}

\maketitle
\ificcvfinal\thispagestyle{empty}\fi

\begin{abstract}
    In this work, we focus on designing a point local aggregation function that yields parameter efficient networks for 3D point cloud semantic segmentation. We explore the idea of using learnable neighbor-to-grid soft assignment in grid-based aggregation functions. Previous methods in literature operate on a predefined geometric grid such as local volume partitions or irregular kernel points. A more general alternative is to allow the network to learn an assignment function that best suits the end task. Since it is learnable, this mapping is allowed to be different per layer instead of being applied uniformly throughout the depth of the network. By endowing the network with the flexibility to learn its own neighbor-to-grid assignment, we arrive at parameter efficient models that achieve state-of-the-art (SOTA) performance on S3DIS with at least 10$\times$ less parameters than the current reigning method. We also demonstrate competitive performance on ScanNet and PartNet compared with much larger SOTA models. 
\end{abstract}
\section{Introduction}
Point cloud semantic segmentation is one of the fundamental tasks of 3D scene understanding. This task provides a holistic view of a scene by grouping points according to their class labels. In an analogous way, 2D semantic segmentation groups pixels into semantically meaningful classes. In image space, Convolutional Neural Networks (CNNs) are currently the de-facto solution to most 2D image understanding tasks including 2D semantic segmentation. Convolutions on 2D images can be done in an optimized and efficient manner thanks to the structured grid of images, unlike point clouds. Extending CNNs to point cloud tasks is challenging, given the intrinsic differences with images. Point sets are unordered and permutation invariant. This is a consequence of how points are localized, mainly by their XYZ coordinates instead of indices in a regular 2D grid seen in images. Also, point clouds are characterized by a generally non-uniform point density distribution over the volume they occupy. For these reasons, discrete kernel convolutions, which are the building blocks of 2D CNNs, cannot be directly applied to point clouds. Efforts exist to project a point cloud into multiple 2D images and then feed them into standard CNNs \cite{mvcnn}. Alternatively, one can voxelize a point cloud into a regular 3D grid that can be processed by a 3D CNN \cite{voxnet}.

\begin{figure}
    \centering
    \includegraphics[trim = 110mm 45mm 110mm 45mm, clip, width=\linewidth]{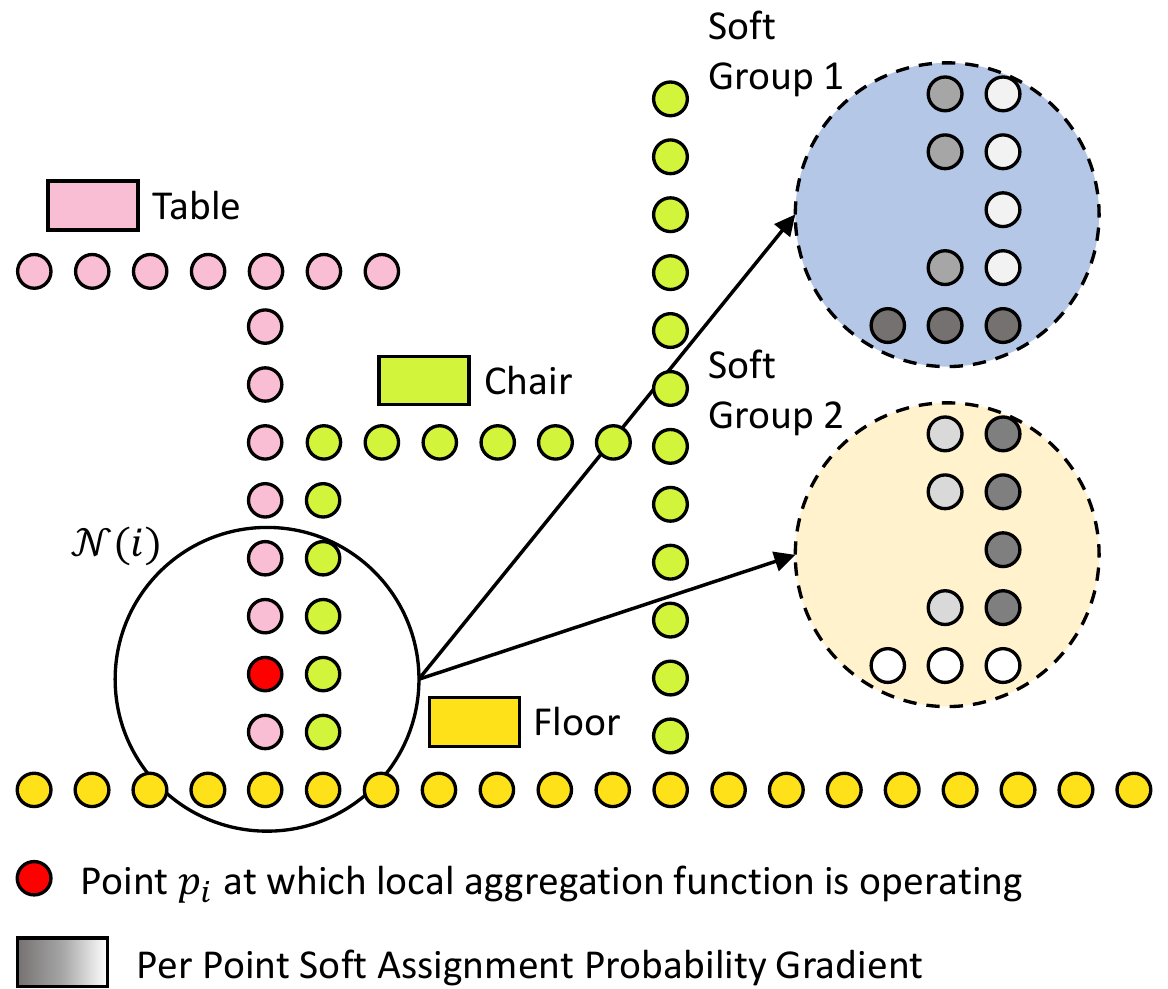}
    \caption{\textbf{Neighbor Point Soft Assignment.} We propose to use learnable soft assignment when aggregating over a neighborhood $\mathcal{N}(i)$ using a grid-based local aggregation function to update the feature of a point $p_i$. This is especially useful when a geometric assignment is not intuitive. The schematic shows 2 groups only, but the assignment is valid for any $S$ number of groups.}
    \label{fig:pullfig}
\end{figure}

\begin{figure*}
    \centering
    \includegraphics[trim = 33mm 58mm 33mm 60mm, clip, width=\textwidth]{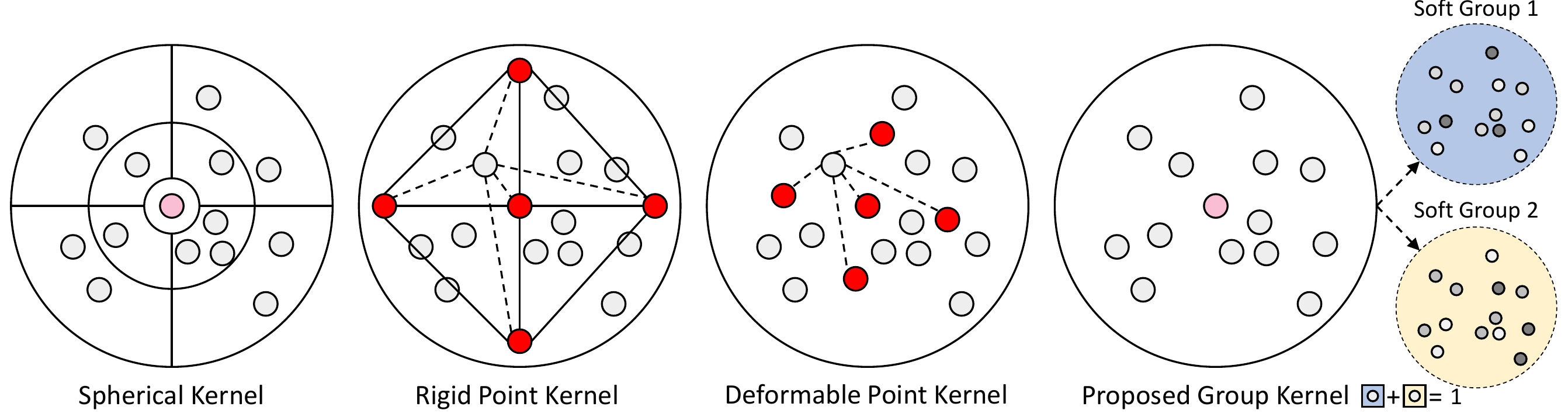}
    \caption{\textbf{Comparison of Spatial Grid Kernels as Opposed to our Non-spatial Defined Kernel.} There exists several spatial grid kernels in literature. Spherical Kernels \cite{sph3d, seggcn} are defined by regular spherical partitions. On the other hand, both rigid and deformable Point Kernels \cite{kpconv} are irregular, since they rely on spatially defined kernel points (red). Unlike rigid point kernels, a deformable kernel adapts its kernel points configuration based on locality. These grid kernels are accompanied by a geometric neighbor-to-grid mapping function. In our case, having a learnable soft assignment function eliminates the need for a spatially defined kernel. Instead, we define groups, each with its own kernel weight. The rightmost figure shows an example of a group kernel with two groups, whereby each neighbor point is endowed with two group membership scores (add up to one) used in aggregation. Clearly, any number of groups can be used.}
    \label{fig:kernels}
\end{figure*}

A more intuitive approach is to devise networks that directly operate on points. These so called point-processing networks are categorized as either global or local neighborhood aggregation methods. As the name suggests, global methods aggregate information from the entire point cloud and fail to capture local contextual details. On the other hand, local neighborhood methods address this problem by defining a point neighborhood, over which information is aggregated (Figure \ref{fig:pullfig}). A typical approach in the local category encompasses grid-based methods, which share a similar flavor to image convolutions. These methods introduce a grid kernel, with cells defined by their spatial grid location and a kernel weight. To operate on point neighborhoods, grid-based methods need to first define the spatial geometry of the grid kernel, as well as the neighbor-to-grid assignment, \ie, which kernel weights interact with each neighbor point. Both these operations are currently pre-defined using a set of geometric heuristics. For example, \cite{sph3d} proposes to use spherical neighborhood partitions as grid cells and assigns kernel weights to neighbor points based on the partition they belong to (refer to Figure \ref{fig:kernels}).

Previous segmentation works mainly focused on enhancing the SOTA performance with little to no focus on efficiency. Efficiency can be defined in regards to inference/training memory, time, power consumption, or network parameter count. Efficiency for point cloud processing is emerging as an important topic with the increased use of LiDAR sensors in embedded systems such as mobile phones and autonomous robots. In this work, we focus on parameter efficiency to guide our design choices.

In this work, we design parameter efficient local aggregation functions for point cloud processing networks. Instead of relying on geometric heuristics, we propose to \textit{learn} the neighbor-to-grid assignment. By learning such an assignment, we do not need to spatially define our grid kernel. Instead, we propose a group-based kernel that consists of groups, to which neighbors are assigned. This kernel enjoys the flexibility of being irregular, unstructured, and non-spatial (refer to Figure \ref{fig:kernels}). Furthermore, neighbor points are assigned to groups in a soft manner, \ie each point belongs to all groups with a learned membership metric. By giving the local aggregation function this flexibility, it can learn an adaptive and a more general neighbor point assignment. This function learns to group neighbor features in a way that best suits the end task of semantic segmentation with no local spatial constraints. Furthermore, the number of parameters used in current SOTA segmentation methods is very large (at least 14M parameters) \cite{kpconv,closerlook3d}, since they rely on increasing network width in order to increase the learning capacity of their models. On the other hand, we empirically show that a learnable assignment improves the expressiveness of the learned features, while using a relatively narrow network resulting in a massive reduction in the number of parameters. We demonstrate the effectiveness of our group-based kernel by formulating a \textbf{S}oft \textbf{A}ssignment based \textbf{L}ocal \textbf{A}ggregation (SALA) function (Figure \ref{fig:kernels}). SALA is integrated into the unified framework proposed by \cite{closerlook3d} for the task of 3D semantic segmentation of indoor scenes and shapes. With this setup, we demonstrate how SALA achieves SOTA performance on S3DIS using at least 10$\times$ less parameters than previous methods. We further show how parameter efficiency translates into networks that have substantially less amount of computations and model memory footprint. We also present very competitive results on ScanNet and PartNet benchmarks, when compared to much bigger models.

\noindent\textbf{Contributions:} \textbf{(i)} We define a novel aggregation called the \textbf{S}oft \textbf{A}ssignment \textbf{L}ocal \textbf{A}ggregation (SALA) function, which is endowed with a learnable and non-spatial grid kernel. SALA offers the flexibility of softly assigning neighborhood points to a grid, it can learn optimal point memberships, and it can enhance the downstream task of point cloud semantic segmentation while using a relatively narrow network. Moreover, SALA is a local aggregation function designed with parameter efficiency at its core. \textbf{(ii)} We run extensive experiments on the challenging S3DIS dataset, where we achieve SOTA performance with at least $10\times$ less parameters. We also show how SALA offers a better trade-off between performance, computation, and model memory footprint compared to other methods. \textbf{(iii)} We demonstrate the efficacy of SALA on other 3D scene and shape part segmentation datasets and we achieve competitive results in ScanNet and PartNet, while using $10\times$ and $3\times$ fewer parameters for scene and shape part segmentation, respectively.

\section{Related Work}
\noindent\textbf{Image- and Voxel-Based Methods.} These methods derive an intermediate representation of point clouds that can be processed by CNNs for 3D point cloud understanding tasks. Image-based methods project a point cloud onto 2D images either using multiple global \cite{mvcnn, deepproj, vmvcnn} or many local tangential view points \cite{tangentconv}. Projecting a point cloud into 2D may result in a very sparse image depending on local point density. Furthermore, local geometric information important for semantic segmentation is compromised, if the view points are not carefully selected \cite{mvf}. In comparison, voxel-based methods leverage the regularity of 3D voxels, on which 3D CNNs can operate \cite{vmvcnn, voxnet}. These methods are generally memory expensive despite being limited to relatively small 3D CNN kernel sizes. Furthermore, a voxel representation uses a fixed size grid throughout a scene, including empty spaces. Alternatively, other methods operate on Octrees \cite{ocnn,octnet} that use an adaptive voxel grid size determined by local point density. Other methods leverage the sparsity of voxel representations by operating on hash tables \cite{submancnn,deepfusion} for efficient feature learning.

\noindent\textbf{Point-Based Methods.} Point processing methods directly operate  on point clouds. They can be classified into global and local methods. PointNet \cite{pointnet} is among the first works for deep learning on un-ordered point clouds. It is a global pointwise method, as it processes neighbor features independently and aggregates information globally by reducing over the entire point cloud through a symmetric aggregation operator. Local methods operate on local neighborhoods instead and can be categorized into three categories:

\noindent\emph{Pointwise-MLP Methods.} PointNet++ \cite{pointnet++} extends PointNet to capture local geometries and contextual relationships by reducing over local spatial neighborhoods instead. Local pointwise-MLP methods such as \cite{pointnet++,dgcnn,deepgcns} treat all neighbors equally and are agnostic to the diversity or redundancy of information in a given neighborhood.

\noindent\emph{Attention-Based Methods.} These methods process neighbor points similar to local pointwise-MLP methods, but replace the simple reduction function by a learnable weighted sum. These weights are generated from a series of transformations on geometric features, such as position \cite{rsconv} and local point density \cite{pointconv}. Alternatively, they use neighbor features to derive attention aggregation weights \cite{gacnet, randlanet}, as in graph attention networks \cite{gat}. The reduction function is still symmetric, unaffected by permutation. By applying an attention on the neighbor features, the local aggregation function would learn to highlight the important neighbors depending on their relevance to learning the end task. Our work is more in line with grid-based methods, which have been empirically shown to be superior to their attention-based counterparts. Instead of learning an attention vector, our method learns to assign each point softly to the different groups of a group-kernel. Therefore every group in our kernel will influence every neighbor point in a given locality.

\noindent\emph{Grid-Based Methods.} These methods attend to neighborhood diversity by projecting a local neighborhood onto a grid kernel. The differentiating factor among these methods is the defined spatial grid and its accompanying neighbor-to-grid mapping function. \cite{sph3d} uses a spherical kernel defined by partitioning a local spherical neighborhood along the spherical coordinates. Neighbor points are then processed by the single kernel weight of the partition, to which they belong. \cite{seggcn} addresses the problem of weight discontinuity at boundary points between partitions by combining the kernel weights of surrounding partitions. \cite{kpconv} introduces kernel point convolutions on local neighborhoods. Kernel points can be regarded as an irregular spatial grid defined by points in a particular configuration. It processes every neighbor point with a weight derived as a linear combination of all kernel weights. The contribution of a kernel weight to a neighbor point is defined by its spatial proximity to the predefined kernel point. \cite{kpconv} also offers a deformable point kernel that learns to shift its initial kernel point configuration per locality. A learned neighbor-to-grid assignment (such as the one proposed in this paper) allows the local aggregation function to explore neighbor interactions beyond spatial similarity heuristics. More importantly, it enables the formation of non-spatial and unstructured kernels, which adapt to each locality. Intuitively, this ability to adapt to different neighborhoods and the lack of spatial kernel definition allows for more expressiveness and discrimination in learned features with a narrower network and, consequently, it significantly reduces the number of needed parameters.
\section{Methodology}
\begin{figure*}[t]
    \begin{center}
    \includegraphics[trim = 15mm 40mm 13mm 40mm, clip, width=\textwidth]{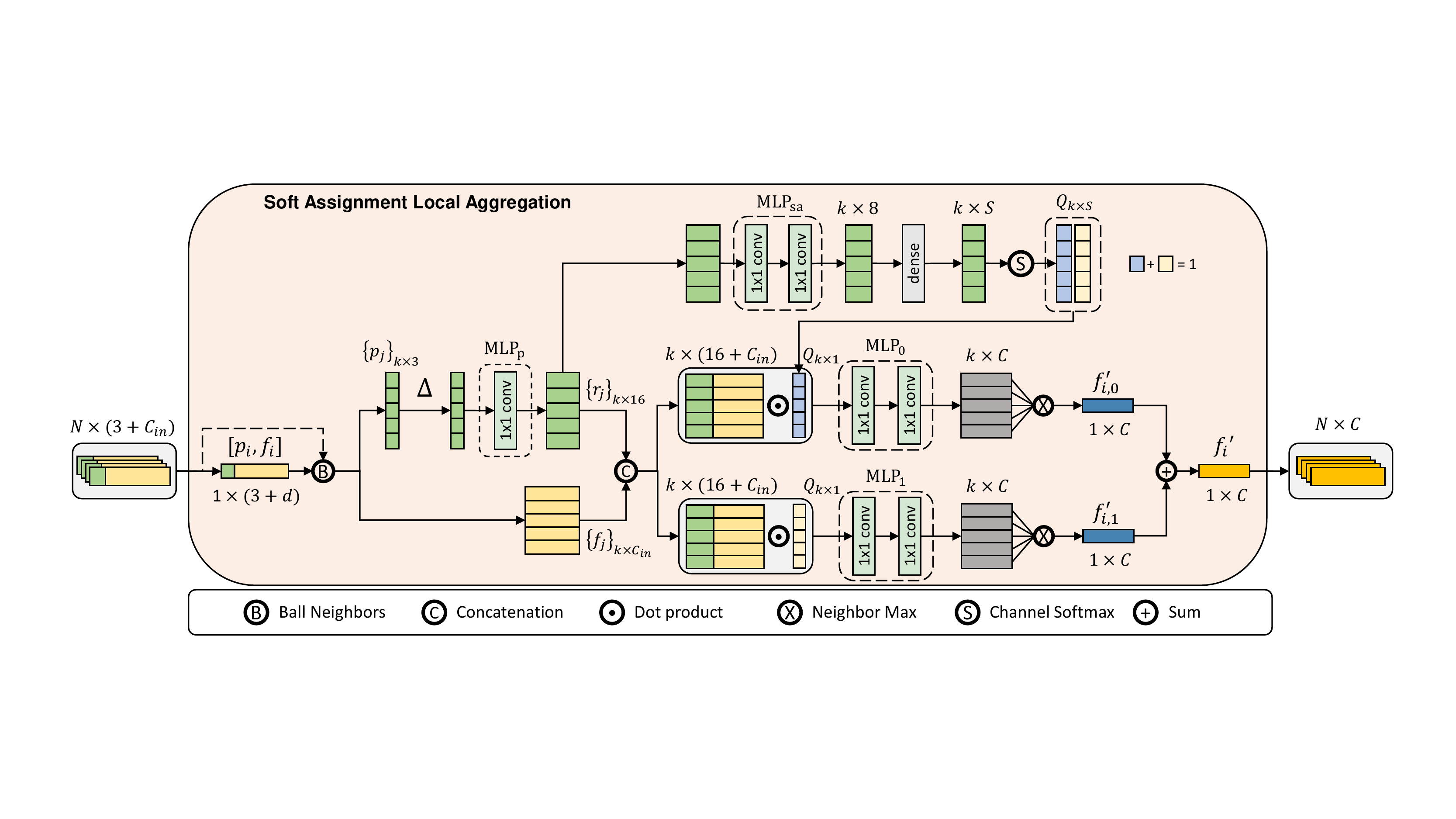}
    \caption{\textbf{SALA: Soft Assignment Local Aggregation.}
    Our proposed local aggregation function leverages learnable soft assignment for grid-based local aggregation functions. SALA can be defined with any number of groups $S$. The schematic figure shows the example with $S=2$ groups only for brevity. The $k$ neighbor points are mapped to the pre-set groups using $\text{MLP}_{sa}$ (\ref{eq:sa}), where each group has its own weight $W_g$ denoted by $\text{MLP}_g$ with $g<S$. Every group branch yields an updated center feature $f'_{i,g}$ (\ref{eq:pergroupfeature}). The final center feature $f'_i$ is obtained by channel-wise sum over updated center features across groups (\ref{eq:featureupdate}). A dense block is a fully connected layer.}
    \label{fig:salablock}
    \end{center}
\end{figure*}

We introduce the SALA function that is designed around parameter efficiency by learning soft assignments of points in grid-based methods instead of relying on geometric heuristics. We show later in the experiments section how models endowed with SALA achieve SOTA or competitive performance, are more efficient in terms of parameters, and are an attractive candidate in the trade-off between compute-efficiency and performance. We also show a comparison of SALA with some of its closest types of local aggregation functions, particularly deformable KPConv \cite{kpconv}, since both can adapt to point localities.

\subsection{Definitions}
Per layer, a point cloud is defined by its 3D coordinates $\mathcal{P}\in \mathbb{R}^{N\times 3}$ and a corresponding set of features $\mathcal{F}\in \mathbb{R}^{N\times C_{in}}$. 
Given a point $p_i \in \mathcal{P}$, a local aggregation function updates its feature $f_i \in \mathcal{F}$ by aggregating information over a local neighborhood $\mathcal{N}(i)$. A grid-based local aggregation function is defined by a kernel of size $S$ and a set of kernel weights $\{W_g\}^{S-1}_{g=0}$ used to encode neighbor features. Previous methods used spatial kernels such as regular spherical partitions or irregular point kernels. An assignment function $h(\cdot)$ maps a neighbor point $j \in \mathcal{N}(i)$ to its corresponding kernel weight. Due to the spatial nature of the kernel, the neighbor-to-grid assignment function $h(\cdot)$ is geometric in nature and based on spatial similarity heuristics. The output of $h(\cdot)$ can either be hard assignment by selecting a single kernel weight \cite{sph3d}, or soft assignment by combining several of them \cite{seggcn, kpconv}. The resulting neighbor encoding weight per grid unit is referred to as $W_{j,g}$ \eqref{eq:generalperunitweight}. The updated feature $f'_{i,g}$ of a  center point per grid unit is then obtained by reducing over the set of its encoded neighbor features using a symmetric operator $R(\cdot)$. The symmetry of $R(\cdot)$ accounts for the permutation invariance property of point clouds \eqref{eq:reduceset}. The final updated center point feature $f'_i$ is obtained by channel-wise sum across the updated center features per grid unit \eqref{eq:sumreduce}. For a grid convolution, $R(\cdot)$ is set to SUM.

\begin{equation} \label{eq:generalperunitweight}
    W_{j,g} = h(p_j,g) W_g
\end{equation}
\begin{equation} \label{eq:reduceset}
    f'_{i,g} = R(\{W_{j,g}f_j\}\mid j\in \mathcal{N}(i))
\end{equation}
\begin{equation} \label{eq:sumreduce}
    f'_i=\sum_{g<S} f'_{i,g}
\end{equation}

In this paper, we propose a learnable soft assignment of neighbors to replace the geometric neighbor-to-grid assignment function $h(\cdot)$ used in previous local grid-based methods. This assignment gives us an advantage over previous grid kernels, as it allows for a non-spatial kernel definition. As a result, we propose to use a group kernel defined by a pre-set number of non-spatial and unstructured groups. A group extends the definition of a point in a point kernel or a partition in a spherical kernel. The proposed learnable soft assignment function regresses the membership probabilities of neighbor points to each group. This function would learn to group neighbor points, whose semantic interaction leads to an improved feature update for the center point. 

\subsection{SALA Formulation}
Operations in SALA are outlined in Figure \ref{fig:salablock}. Similar to grid-based methods, every group has its own learnable weight $W_g$ defined by $\text{MLP}_g$ with $g<S$ and used to transform the input neighbor feature. The soft assignment function $h(\cdot)$ regresses membership probabilities per neighbor point to each group using $\text{MLP}_{sa}$. It operates on the encoded relative positions $r_j$ \eqref{eq:lift} of a center point $p_i$ and a neighbor point $p_j$ for $j\in \mathcal{N}(i)$. The output of $h(r_j)$ is a probability vector of dimension equal to the number of groups: $h(r_j)\in \mathbb{R}^{1\times S}$ \eqref{eq:sa} . Then, $h(r_j,g)$ corresponds to the probability of a neighbor $j \in \mathcal{N}(i)$ to belong to group $g$. The probability axioms are enforced by a Softmax applied along the group dimension. The stacked soft assignment probabilities of all $k$ neighbors to all $S$ groups are denoted as $Q\in \mathbb{R}^{k\times S}$.
\begin{equation}\label{eq:lift}
    r_{j}=\text{MLP}_p\left(p_j-p_i\right)
\end{equation}
\begin{equation} \label{eq:sa}
    h(r_j)=\text{Softmax}\left(\text{MLP}_{sa}(r_j)\right)
\end{equation}

\noindent A neighbor processing weight is obtained per-group $W_{j,g}$ by scaling the group kernel weight $W_g$ with the regressed soft assignment probabilities \eqref{eq:pergroupweight}. This scaling operation resembles the contribution of a neighbor feature $f_j$ to the per-group updated center feature $f'_{i,g}$. The feature used for local aggregation is denoted by $f^{*}_j$ (\ref{eq:salainputs}), which incorporates the position encoding element $r_j$. Every group branch yields its own version of the updated center point feature by reducing over the set of per-group encoded neighbor features \eqref{eq:pergroupfeature}. The reducing function $R(\cdot)$ of choice is MAX. The final updated center feature $f'_i$ is obtained by reducing over the updated center feature obtained per-group.

\begin{equation} \label{eq:pergroupweight}
    W_{j,g}=h(p_j,g)W_g
\end{equation}
\begin{equation} \label{eq:salainputs}
    f^*_{j}=\text{concat}\left(r_j, f_j\right)
\end{equation}
\begin{equation} \label{eq:pergroupfeature}
    f'_{i,g}=\mathbf{MAX}(\{W_{j,g} f^*_{j}\}\mid j \in \mathcal{N}(i))
\end{equation}
\begin{equation} \label{eq:featureupdate}
    f'_i=\sum_{g<S} f'_{i,g}
\end{equation}

\noindent Note that when $S=1$, the proposed aggregation function simplifies to a pointwise-MLP local aggregation function. In this case, $Q \in \mathbb{R}^{k\times 1}$ becomes a vector of all ones due to the softmax operator.

\begin{figure*}[htb]
    \centering
    \includegraphics[trim = 30mm 50mm 30mm 65mm, clip, width=\linewidth]{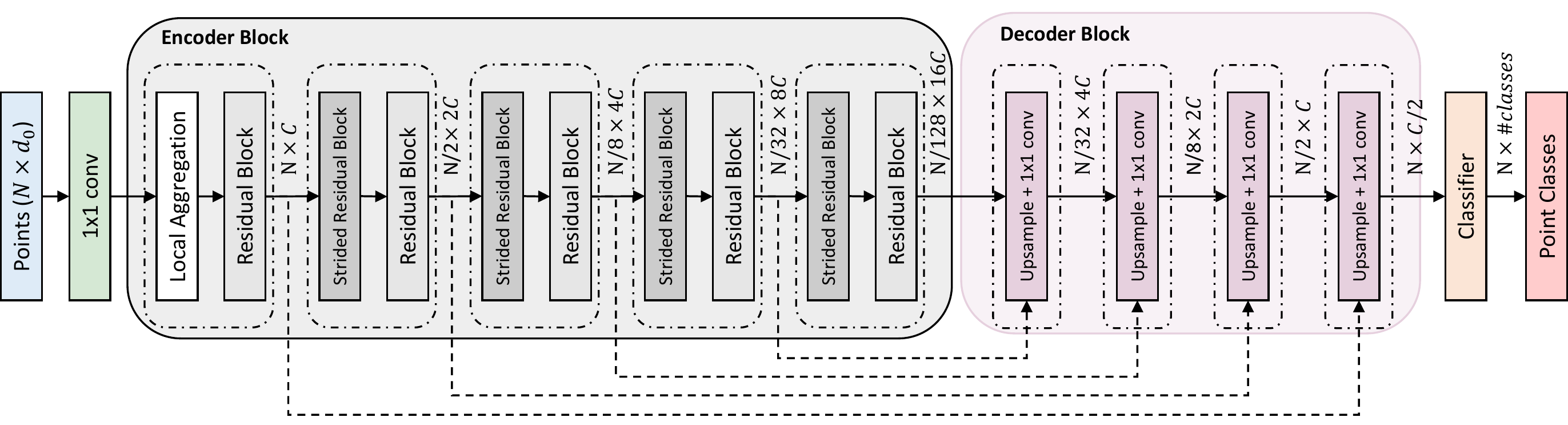}
    \caption{\textbf{Architecture.} Unified architecture framework from \cite{closerlook3d}. Both Residual and Strided Residual Blocks have local aggregation functions within. The strided residual block acts as a pooling operator, in which point cloud resolution is reduced using grid-subsampling. The grid size is doubled going deeper in the encoder. $C$ refers to the channel output of the first encoding layer.}
    \label{fig:arch}
    \vspace{-2mm}
\end{figure*}

\subsection{Comparison with Deformable Point Kernel}
For the purpose of comparing SALA to the deformable KPConv~\cite{kpconv} kernel, we introduce a SALA variant by using SUM function instead of MAX \eqref{eq:pergroupfeature} to reduce over neighbor features per-group. The resulting feature update can be written as follows:
\begin{equation}\label{eq:sumsala}
    f_i'= \sum_{j<k} \sum_{g<S} h(p_j, g) W_g f^*_{j}
\end{equation}

The kernel in KPConv\cite{kpconv} is defined by a set of $S$ kernel points $\{\tilde{x}_{g}\}^{S-1}_{g=0}$ in 3D space within a sphere neighborhood, each with a kernel weight $W_{g}$. Starting with an initialized kernel point configuration, a rigid point kernel maintains the initial spatial configuration throughout all localities. On the other hand, a deformable kernel learns to shift its kernel points by $\Delta_{p_i}$ depending on the center point location $p_i$, where the convolution is being carried out. Point kernel refers to the neighbor-to-grid assignment function $h(\cdot)$ as a kernel point influence function (\ref{eq:kernelinfluence}). It is defined by the normalized distance between a centered neighbor point $r_j$ and a kernel point $\tilde{x}_g$.
\begin{equation}
    r_{j}=p_j-p_i
\end{equation}
\begin{equation}\label{eq:kernelinfluence}
    h(r_j, \tilde{x}_{g})=\mathbf{MAX}\left(0, 1-\frac{\|r_j - \tilde{x}_{g}\|_2}{\sigma}\right)
\end{equation}
\begin{equation}\label{eq:kpconv}
    f_i'=\sum_{j<k} \sum_{g<S} h(p_j, \tilde{x}_{g}+\Delta_{p_i}) W_{g}  f_j
\end{equation}
By comparing equations (\ref{eq:sumsala}) and (\ref{eq:kpconv}), one observes that the group kernel in SALA shares some similarities with the deformable point kernel in \cite{kpconv}. Instead of learning shifts in kernel points, we learn to assign neighbors to non-spatial groups using a learnable assignment function $h(\cdot)$ (\ref{eq:sa}). Similar to a deformable kernel, our learnable function adapts per locality. An additional advantage of SALA over previous methods is that it can learn a different mapping function per layer. The learned assignment will therefore be driven by the performance on the end task.

\subsection{Architecture}
Recently, \cite{closerlook3d} showed that different types of local aggregation functions can reach comparable SOTA performance on several datasets. It uses a similar architecture as proposed in \cite{kpconv}, as well as similar data preprocessing, training, and testing strategies. For a fair and consistent comparison, our proposed local aggregation function is tested within the same architecture and framework as \cite{kpconv,closerlook3d}. The architecture under study is a U-Net inspired architecture consisting of encoder and decoder blocks. An illustration of the architecture is depicted in Figure \ref{fig:arch}.

\noindent \textbf{Encoder Block}: The architecture consists of five encoding blocks comprised of residual and strided residual blocks, each with $1\times1$ convolution operators and a local aggregation function within. The encoder block operates on hierarchical point cloud resolutions to incorporate contextual information at multiple scales. Going from one encoder layer to the other, the point cloud resolution is reduced using point cloud subsampling methods. In order to account for the varying density problem commonly encountered in point clouds, \cite{kpconv} proposes to use grid-subsampling to reduce the resolution while maintaining a uniform point cloud density throughout. The grid size is doubled going from an encoder layer to the subsequent one. The strided residual block acts as a graph pooling operation using a local aggregation block. 
 
Neighbors are queried using a ball neighborhood that retrieves more representative neighbors as compared to the density sensitive k-nearest-neighbors \cite{ballknn}. The ball neighborhood radius is doubled going deeper in the encoder to account for the increasing sparsity of the point cloud due to downsampling. The features of the lowest resolution at the end of the encoder block are propagated to the full cloud resolution by the decoder block.

\noindent \textbf{Decoder Block}: It consists of nearest neighbor interpolation and $1\times1$ convolution operations. Nearest neighbor interpolation copies the features of points at low resolution to their neighbors in the subsequent decoder point cloud at higher resolution. Skip connections inject encoder features at the same resolution as upsampled decoder features. The two are concatenated and aggregated by a $1\times1$ convolution operator. \cite{closerlook3d} introduces the parameter $C$ as the output channel dimension of the first local aggregation function as a way to control the size of the model.
\section{Experiments}
To demonstrate the effectiveness of our proposed local aggregation function, we evaluate models under the same settings proposed in \cite{kpconv, closerlook3d} for the task of indoor scene semantic segmentation and \cite{closerlook3d} for shape part segmentation.
\subsection{Scene Semantic Segmentation on S3DIS}
\begin{table}[t]
    \small
    \tabcolsep=0.13cm
    \begin{center}
        \begin{tabular}{l|c c c}
            \hline
            Method & S3DIS A5 & S3DIS 6F & ScanNet \\
            \hline
            PointNet \cite{pointnet} & 41.1 & - & -\\
            PointNet++ \cite{pointnet++} & - & - &33.9\\ 
            PointCNN \cite{pointcnn} & 57.5 & 65.4 & 45.8\\
            SSP+SPG \cite{sspspg} & 61.7 & 68.4 & - \\
            SPH3D-GCN \cite{sph3d}& 59.5 & 68.9 & 61.0 \\
            SegGCN \cite{seggcn} & 63.6 & 68.5 & 58.9 \\
            PointASNL \cite{pointasnl} & 62.6 & 68.7 & 66.6 \\
            RandLANet \cite{randlanet} & - & 70.0 & - \\
            Pseudo Grid (18.5M) \cite{closerlook3d} & 65.9 & - & - \\
            PointwiseMLP (25.5M) \cite{closerlook3d} & 66.2 & - & - \\
            Adapt Weights (18.4M) \cite{closerlook3d} & 66.5 & - & - \\
            PosPool (18.5M)\cite{closerlook3d} & 66.5 & - & - \\
            KPConv (r) (14.1M)\cite{kpconv}& 65.4 & 69.6 & \textbf{68.6} \\
            KPConv (d) (14.9M)\cite{kpconv}& 67.1 & 70.6 & 68.4 \\
            \hline
            \hline
            2 groups - SALANet (1.6M) & \textbf{67.6} & \textbf{72.5} & 67.0\\
            \hline
        \end{tabular}
        \vspace{1pt}
        \caption{\textbf{State-of-the-art mIoU comparison on S3DIS and ScanNet}. SALANet with $S=2$ groups achieves SOTA on S3DIS test Area $5$ (A5) and $6$-fold-cross-validation (6F), and competitive performance on ScanNet test set in comparison with much bigger models. Only point processing methods are considered for ScanNet. $\mathbf{M}$ corresponds to million parameters.}\label{table:semseg}
        \end{center}
        \vspace{-5mm}
\end{table}

We conduct experiments on the challenging indoor scene semantic segmentation dataset S3DIS \cite{s3dis}. 
The output feature dimension of the first local aggregation is set to $C=36$ as proposed for the small model variants in \cite{closerlook3d}. On S3DIS, ZRGB1 point features are used as initial inputs to the model following \cite{kpconv,closerlook3d}. The concatenated ``1" feature is meant to account for all black RGB points. The downsampling base grid size is set to $0.04$ meters for both datasets. S3DIS \cite{s3dis} consists of six large areas of indoor scenes scanned using a Matterport device. Every point is represented by its XYZ coordinates, RGB color features, and a class label that takes one of thirteen different semantic classes. Following common practice, we report mean intersection over union (mIoU) on both Area 5 and 6-fold-cross-validation benchmarks to demonstrate the generalization of our proposed aggregation function. 

\begin{figure*}[htb]
    \centering
    \includegraphics[trim = 2mm 25mm 2mm 25mm, clip, width=\linewidth]{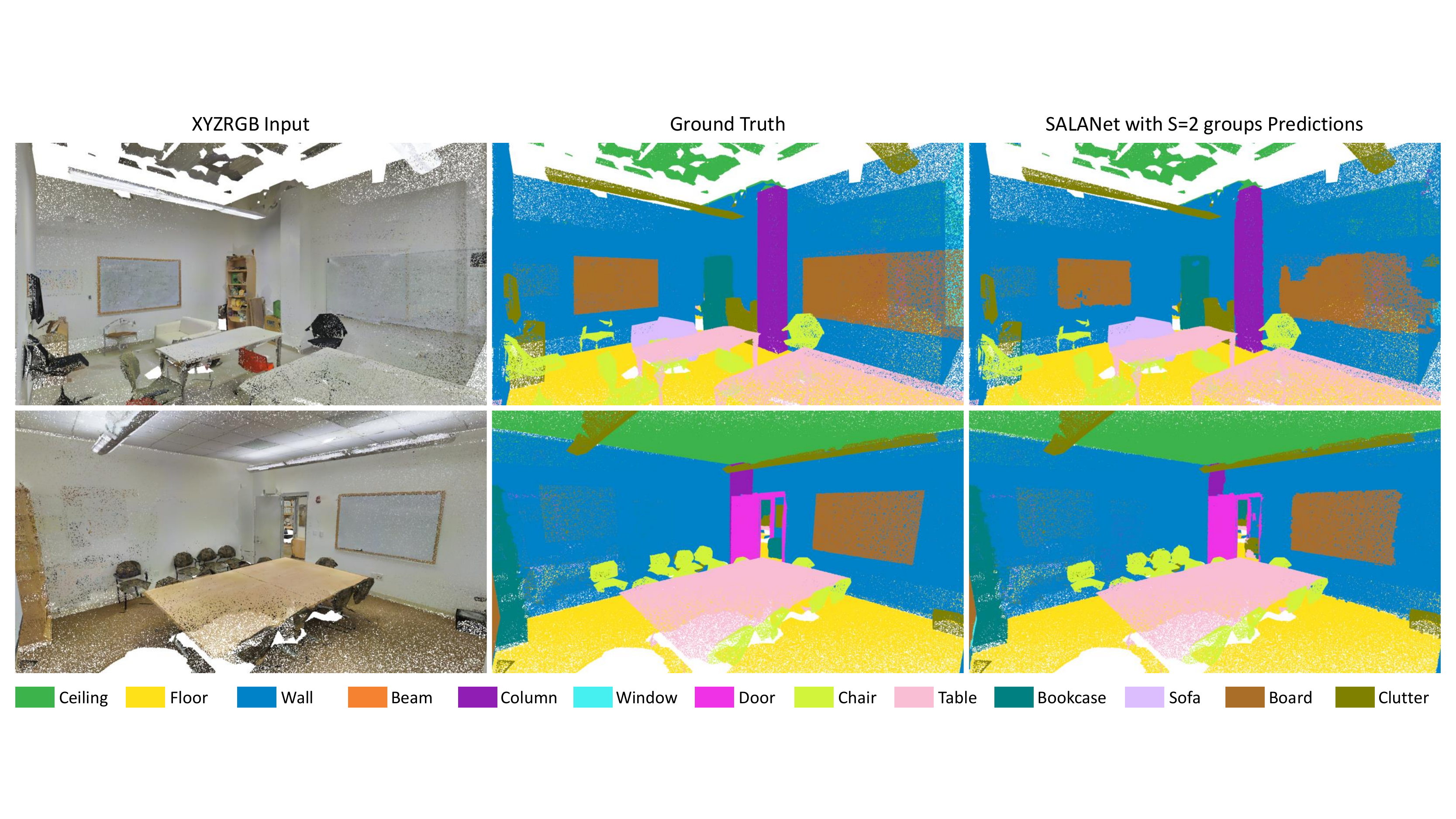}
    \caption{\textbf{Qualitative Results.} Sample qualitative results of SALANet with $S=2$ groups on S3DIS Area 5. SALANet attends to easily confused classes such as wall/column, wall/board and wall/door that are all semantically similar.}
    \label{fig:qualitative}
    \vspace{-2mm}
\end{figure*}

In training, random spheres are sampled with a radius $r=2$ meters. In testing, overlapping spheres are sampled regularly, so that every point is inferred multiple times in different contexts as a voting scheme, following \cite{kpconv, closerlook3d}. The multiple predictions are then averaged and a point label is assigned. SALANet refers to the model configured with SALA as the local aggregation function. Results of SALANet on 3D scene semantic segmentation are summarized in Table \ref{table:semseg}. On S3DIS Area 5, SALANet with $S=2$ groups reaches $67.6$ mIoU defining the new SOTA on this benchmark. Similarly, it reaches $72.5$ mIoU on the 6-fold-cross-validation benchmark outperforming the current SOTA method by close to $2$ mIoU points, while using $10\times$ less parameters. The bigger performance boost in the 6-fold-cross-validation demonstrates the generalization capability of our proposed local aggregation function. Sample qualitative results on S3DIS Area 5 are shown in Figure \ref{fig:qualitative}. We explore even smaller SALANet model later in Section \ref{sec:emacs}.

\subsubsection{Compute Efficiency and Performance Trade-off} \label{sec:emacs}
\begin{table}[t]
    \small
    \tabcolsep=0.07cm
    \begin{center}
        \begin{tabular}{l|c|c|c}
            \hline
            Method & mIoU & GMACS & Memory (MB)\\
            \hline
            Point-wise MLP (25.5M)\cite{closerlook3d} & 66.2  & 73.0 &  203.9 \\
            Adapt Weights (18.4M)\cite{closerlook3d}  & 66.5 & 7.3 & 147.4 \\
            PosPool (18.1M)\cite{closerlook3d} & 66.5 & 6.8 & 147.2 \\
            \hline
            \hline
            2 groups-SALANet (0.41M) & 64.9 & 4.6 & 3.8 \\
            2 groups-SALANet$^*$ (1.6M) & 67.6 & 12.9 & 13.5 \\
            \hline
        \end{tabular}
        \vspace{1pt}
        \caption{\textbf{GMACS and Storage Memory Footprints}. We take record of GMACS required for a forward pass of 15000 points for several methods in \cite{closerlook3d}. We also record the storage memory footprint for each model and compare with SALANet. mIoU on S3DIS Area 5 is also reported for reference. (*) denotes the configuration of choice. 
        $\mathbf{M}$ corresponds to million parameters.}\label{table:macs}
        \end{center}
        \vspace{-5mm}
\end{table}
In Table \ref{table:macs}, instead of relying on heuristics to compute the GFLOPS performed by different models, we take record of GMACs (Giga-multiply-accumulate operations) needed for a forward pass over the entire network. We also report each model's memory footprint, the memory needed to store the weights of a trained model. We evaluate GMACS using 15000 points under the same settings for several models in \cite{closerlook3d} and compare them to SALANet. We also explore an even smaller SALANet model with $C=18$ that achieves $64.9$ mIoU on S3DIS Area 5, while only using 0.41 million parameters. The smaller variant's performance is on par with the rigid kernel KPConv model that reaches 65.4 mIoU and uses more than 30$\times$ more parameters. This clearly demonstrates the expressiveness of SALANet models.

Although the chosen setting of SALANet requires more GMACS (12.9 GMACS) compared to larger models in \cite{closerlook3d}, it comes with at least 10$\times$ reduction in memory footprint. In comparison, the smaller SALANet model comes with fewer GMACS (4.6 GMACS) and a 30$\times$ reduction in memory footprint with an acceptable performance compromise. This makes SALANet desirable for embedded system applications, where storage memory and GMACS are constraints.
 
\subsubsection{Ablations on S3DIS Area 5}
\begin{table}[t]
    \small
    \tabcolsep=0.3cm
    \begin{center}
        \begin{tabular}{l|c}
            \hline
            Ablations & S3DIS A5 (mIoU)\\
            \hline
            $R(\cdot)$ set to SUM & 61.6\\
            No Positional Encoding & 63.5\\
            \hline
            \hline
            Hard Assignment & 64.0\\
            No Assignment & 64.2\\
            \hline
            \hline
            2 groups - SALANet (C=18, 0.41M) & 64.9\\
            2 groups - SALANet$^*$ (C=36, 1.6M) & 67.6\\
            \hline
        \end{tabular}
        \vspace{3pt}
        \caption{\textbf{Ablation of SALANet on S3DIS Area $5$}. We report results of SALANet model variants using SALA with SUM as group reduction $R$ function and removing positional encoding from local aggregation. We further study the contribution of soft assignment by disregarding the regressed soft assignment scores, and thresholding soft assignment scores to hard assignment.  (*) denotes the configuration of choice. $\mathbf{M}$ corresponds to million parameters. 
        }\label{table:ablation2}
        \end{center}
\end{table}
The adopted setting of SALANet uses $S=2$ groups, MAX as the group reduction function $R$  (\ref{eq:pergroupfeature}), $f^*_j=[r_j, f_j]$ (\ref{eq:salainputs}) as input for local aggregation, and $C=36$ as the width of the first encoder block. The number of groups is fixed to $S=2$ for parameter efficiency, as well as being the minimum number of groups for which the soft assignment is meaningful. We conduct an ablation study on the width of SALANet, importance of soft assignment scores in local aggregation, choice of group reduction function, and importance of positional encoding as summarized in Table \ref{table:ablation2}. 

\noindent\textbf{Group Reduction Function.} $R(.)$ is set to SUM for grid convolution methods such as \cite{kpconv,sph3d,seggcn}. In SALA, we find that replacing $R(.)$ with SUM significantly deteriorates the performance of the resulting SALANet model by 6 mIoU points compared to the setting of choice (MAX).

\noindent\textbf{Positional Encoding.} Removing the positional encoding from the local aggregation feature yields $f^*_j=[f_j]$ and results in 4 mIoU points drop compared to the setting of choice. We believe the benefit of using lifted relative position $r_j$ as positional encoding decouples the neighbor search space from point feature learning. Moreover, using the lifted relative position will allow gradients a better flow back to the learnable assignment function to further improve the soft assignment process.

\noindent\textbf{Importance of Soft Assignment.} Two experiments were conducted to assess the importance of soft assignment scores in local feature aggregation. \emph{(i)} Binarizing soft assignment scores by thresholding them at 0.5. This results in discontinuity in the neighbor processing weight per-group $W_{j,g}$ (\ref{eq:pergroupweight}), especially when two points are spatially close but favor two different groups. The performance of this hard assignment model drops by $3.6$ mIoU points. \emph{(ii)} Disregarding soft assignment in local aggregation by setting all assignment scores to one results in a drop in performance of 3.4 mIoU points.

\subsection{Scene Semantic Segmentation on ScanNet}
\vspace{-2mm}
\noindent ScanNet \cite{scannet} consists of $1513$ training scenes scanned using RGB-D cameras. Every point is represented by its XYZ coordinate, RGB color features, and a class label that takes one of twenty different semantic classes. The train/val split is set as suggested by the benchmark. The test set comprises $100$ scenes. The reported mIoU metric is computed on the test set, as reported on the ScanNet benchmarks hosting website. On ScanNet, RGB1 features are used as in \cite{kpconv}, the rest of training and inference details are carried out on ScanNet exactly as described for S3DIS. On ScanNet, SALANet reaches $67$ mIoU with $9\times$ less parameters compared to the much more complex rigid kernel of KPConv \cite{kpconv} (refer to Table \ref{table:semseg}). This demonstrates the accuracy-to-efficiency balance of our proposed SALA function, when compared to other SOTA methods.

\subsection{3D Shape Part Segmentation}
\begin{table}[t]
    \small
    \tabcolsep=0.2cm
    \begin{center}
        \begin{tabular}{l|c}
            \hline
            Method & PartNet (mpIoU)\\
            \hline
            ResGCN28 \cite{deepgcnsext} & 45.1 \\
            PointCNN (4.4M)\cite{pointcnn} & 46.5 \\
            \hline
            PointwiseMLP (C=36, 1.6M) \cite{closerlook3d}& 47 \\
            Pseudo Grid (C=36, 1.2M) \cite{closerlook3d} & 45.2 \\
            Adapt Weights (C=36, 1.2M) \cite{closerlook3d}& 46.1 \\
            PosPool* (C=36, 1.1M) \cite{closerlook3d} & 47.2 \\
            \hline
            PointwiseMLP (C=144, 25.6M) \cite{closerlook3d}& 51.2 \\
            Pseudo Grid (C=144, 18.5M) \cite{closerlook3d}& 53 \\
            Adapt Weights (C=144, 18.5M) \cite{closerlook3d}& 53.5 \\
            PosPool* (C=144, 18.5M) \cite{closerlook3d}& \textbf{53.8} \\ 
            \hline
            \hline
            2 groups - SALANet (C=36, 1.6M) & 49.4\\
            2 groups - SALANet (C=72, 6.4M) & 52.4\\
            \hline
        \end{tabular}
        \vspace{3pt}
        \caption{\textbf{Shape Part Segmentation mpIoU on PartNet}. SALANet with $S=2$ groups and $C=72$ achieves competitive performance on PartNet (level 3) test set in comparison with much bigger models. SALANet is trained and tested under the same settings in \cite{closerlook3d}. $\mathbf{M}$ corresponds to million parameters.}\label{table:partnet}
        \end{center}
        \vspace{-5mm}
\end{table}
\vspace{-2mm}
\noindent We conduct further experiments on PartNet for the task of shape part segmentation. PartNet~\cite{partnet} is a large-scale dataset for part segmentation consisting of synthetic 3D CAD models with $24$ shape categories, each with a different set of assembled parts. We consider the most fine-grained segmentation (level-3) task on this dataset, which involves $17$ shape categories. We follow \cite{closerlook3d} by training on all categories at once instead of training different shapes independently. This is done by using the same backbone and an independent classifier head per shape category. We use XYZ1 features as model inputs and a base grid size of $0.02$ meters. We follow the official train/val split and report mean part IoU (mpIoU) on the test set that comprises $20\%$ of the entire dataset. Results of SALANet on 3D shape part segmentation are summarized in Table \ref{table:partnet}.
On PartNet, we test SALANet with $C=36$ and $S=2$ groups and report $49.4$ mpIoU on the test set. This small model variant outperforms all small model variants  proposed in \cite{closerlook3d} (\ie PointwiseMLP, Pseudo Grid, and PosPool with $C=36$). To demonstrate the effectiveness of our aggregation function, we double the output channel size to $C=72$ and show that our SALANet with $S=2$ groups can reach a very competitive performance with almost $3\times$ times less parameters than the large model variants in \cite{closerlook3d} with $C=144$. The comparison is especially valuable with the Pseudo Grid method, which uses point kernel aggregation \cite{kpconv} with a geometric neighbor-to-grid assignment function. We note that further increasing $C$ compromises our parameter efficiency claim.
\vspace{-3mm}
\section{Conclusion}
\vspace{-2mm}
In this work, we presented SALA as a local aggregation function designed around parameter efficiency. It is a grid-based local aggregation function that leverages a learnable neighbor-to-grid assignment. Instead of relying on a spatial grid, our local aggregation function softly assigns neighbor points to a fixed number of irregular, unstructured, and non-spatial groups. We integrate our local aggregation into the unified framework architecture of \cite{closerlook3d} (denoted as SALANet), which is designed for 3D semantic segmentation. We demonstrate the impact of a learnable mapping by achieving SOTA results on S3DIS and competitive performance on ScanNet and PartNet, while using a fraction of the parameters used in SOTA methods. The learned assignment has the flexibility to adapt per layer and locality, improving the expressiveness of the learned features, and it is driven by the performance on the end task of semantic segmentation.
{\small
\bibliographystyle{ieee_fullname}
\bibliography{egbib}
}
\clearpage
\appendix
\noindent \Large \textbf{Supplementary Material}
\normalsize
\section{Architecture Blocks}
\begin{figure*}[htb]
    \centering
    \includegraphics[trim = 30mm 30mm 30mm 5mm, clip, width=\linewidth]{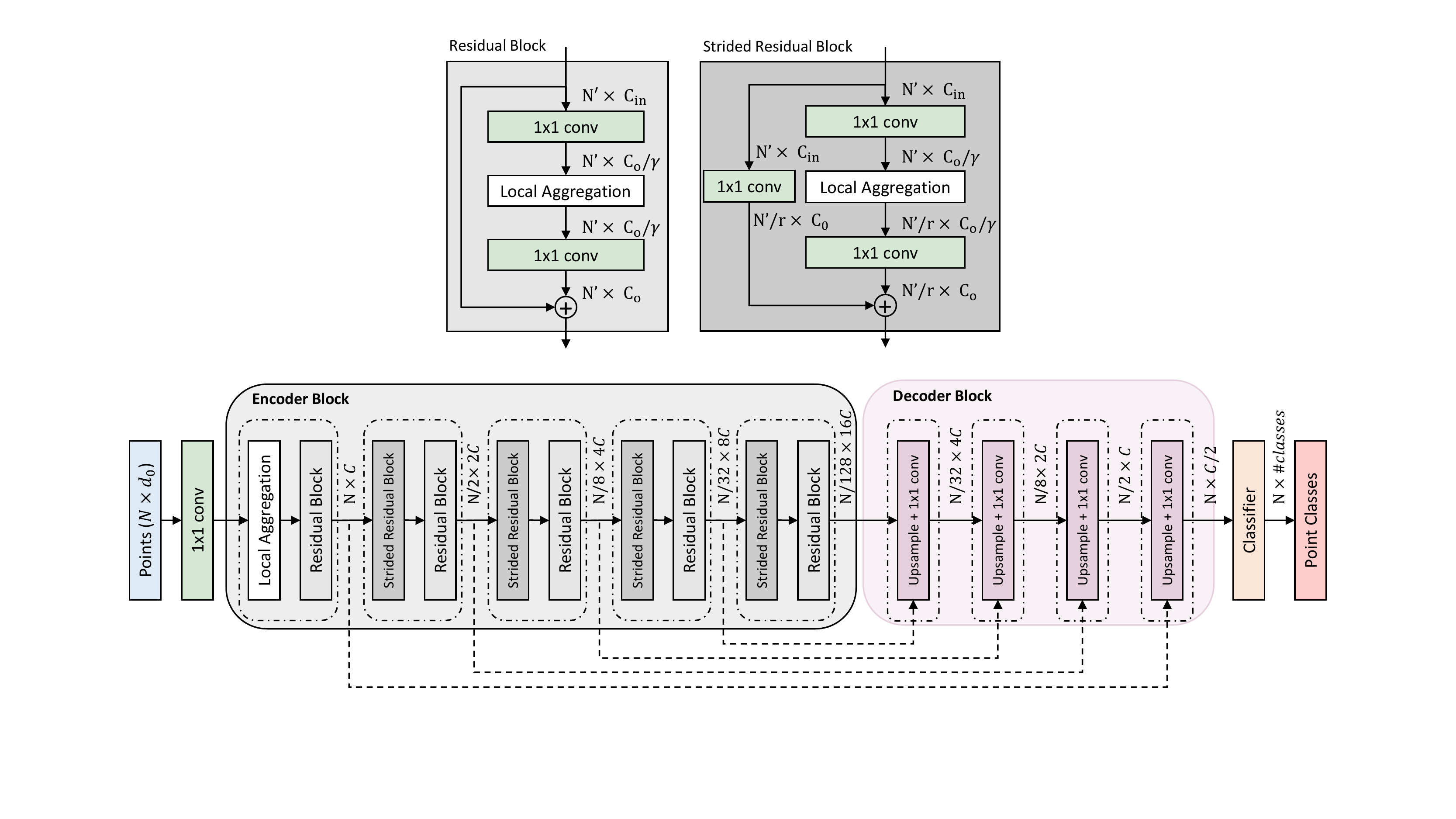}
    \caption{\textbf{Architecture.} Unified architecture framework from \cite{closerlook3d}.
    Both Residual and Strided Residual Blocks have local aggregation functions within. The strided residual blocks acts as a pooling operator in which point cloud resolution is reduced using grid-subsampling. The grid size is doubled going deeper in the encoder. 
    $C$ refers to channel output of first encoding layer. $\gamma=2$ is the residual bottlneck ratio. $r$ refers to the encoder point cloud downsampling ratio.}
    \label{fig:archblocks}
    \vspace{-2mm}
\end{figure*}

Residual and Strided Residual architecture blocks are detailed in Figure \ref{fig:archblocks} as in \cite{closerlook3d}.
\section{Training and Inference Settings}
We evaluate Soft Assignment Local Aggregation (SALA) function in the unified framework proposed in \cite{closerlook3d}. This includes data pre-processing, augmentation, architecture (see Figure \ref{fig:archblocks}), and training and inference schemes. Models are trained with cross-entropy loss and $L_2$ regularizer on learned weights.

\subsection{S3DIS}
The initial inputs to the model are lifted ZRGB1 features ($d_0=5$). Following \cite{closerlook3d,kpconv}, the data is augmented with random rotations along the z-axis, random scaling (ratios 0.7 to 1.3), random color drop, and additive Gaussian point perturbations with standard deviation $\delta=0.001$. The reported mIoU is a result of voting scheme employed in \cite{closerlook3d,kpconv}. A point cloud is divided using regular overlapping spheres. The logits of a point seen in different spheres are then averaged before Softmax and label assignment.

\subsection{ScanNet}
The initial inputs to the model are lifted RGB1 features ($d_0=4$). Following \cite{kpconv}, the data is augmented with random rotations along the z-axis, random scaling (ratios 0.9 to 1.1), random color drop, and additive Gaussian point perturbations with standard deviation $\delta=0.001$. The reported mIoU is as reported on ScanNet benchmarks website. Predictions on the test-set are collected using the same voting scheme discussed previously as in \cite{kpconv}.

\subsection{PartNet}
The initial inputs to the model are lifted XYZ1 features ($d_0=4$). Following \cite{closerlook3d}, the data is augmented with random scaling (ratios 0.8 to 1.2), and additive Gaussian point perturbations with standard deviation $\delta=0.001$. The reported mIoU is a result of voting scheme where logits are averaged over prediction on augmented test-set as in \cite{closerlook3d}. 

\section{Quantitative and Qualitative Results}
\subsection{S3DIS}
\begin{figure*}[htb]
    \centering
    \includegraphics[trim = 2mm 35mm 2mm 30mm, clip, width=\linewidth]{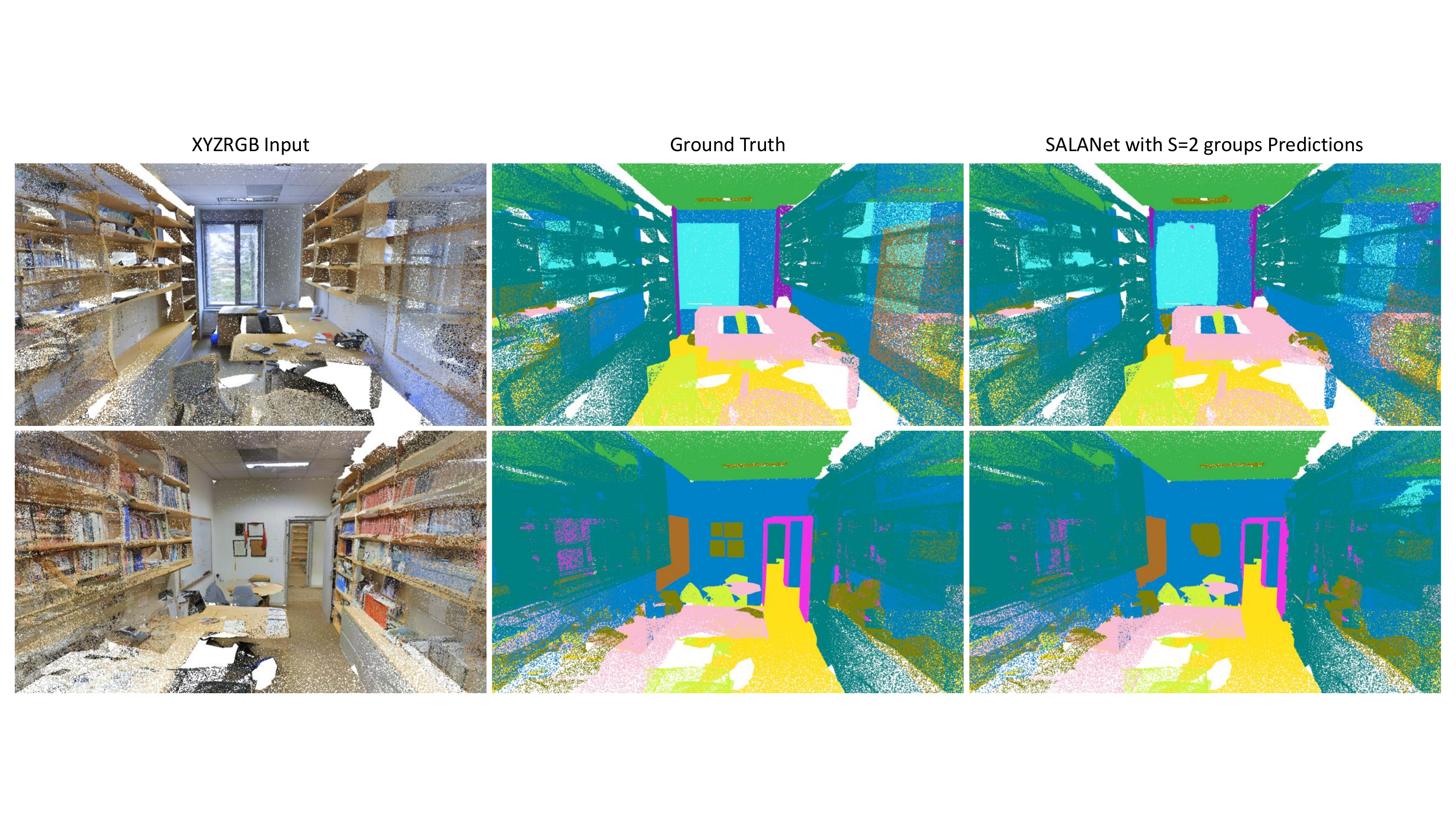}
    \includegraphics[trim = 2mm 30mm 2mm 21mm, clip, width=\linewidth]{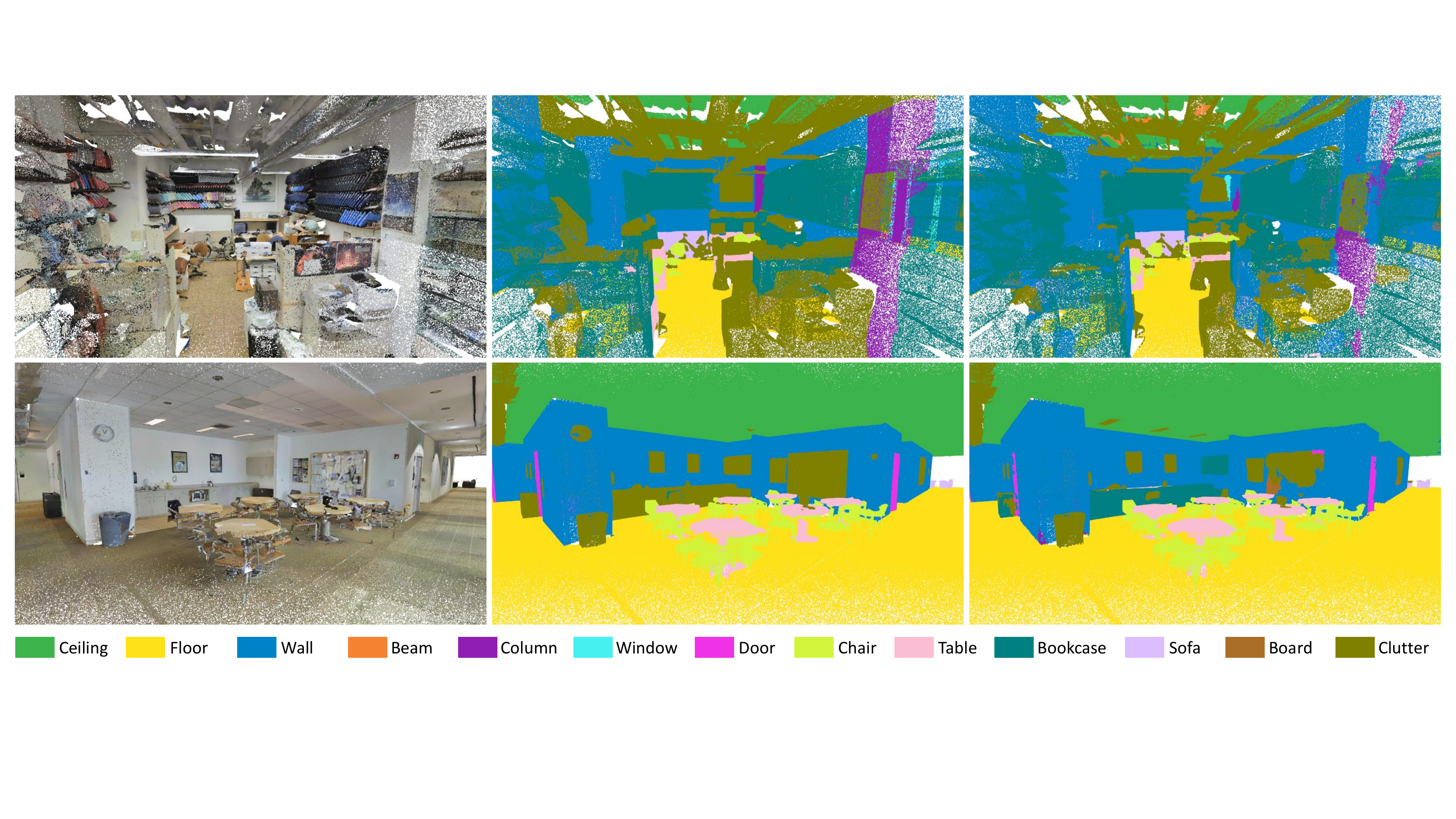}
    \caption{\textbf{Qualitative Results on S3DIS.} Sample qualitative results of SALANet with $S=2$ groups on S3DIS Area 5.}
    \label{fig:supquals3dis}
    \vspace{-2mm}
\end{figure*}

Tables \ref{table:sups3disa5} and \ref{table:sups3dis6fold} show the per class IoU on S3DIS Area 5 and 6-fold-cross-validation benchmarks respectively. On both benchmarks, SALANet with $S=2$ groups achieves state-of-the-art (SOTA) performance. Sample qualitative results are showcased in Figure \ref{fig:supquals3dis}.

\subsection{ScanNet}
\begin{figure*}[htb]
    \centering
    \includegraphics[trim = 2mm 5mm 2mm 5mm, clip, width=\linewidth]{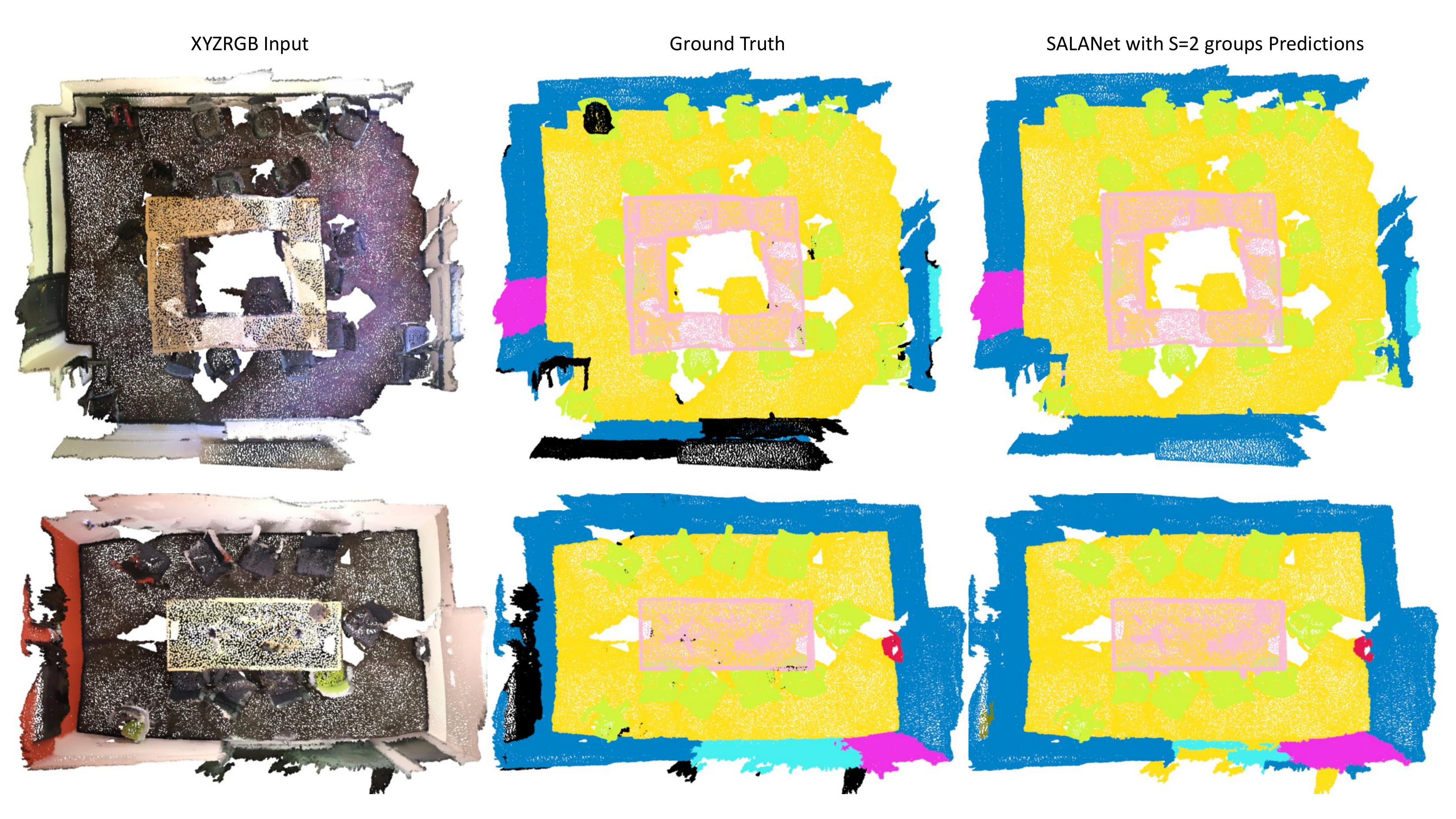}
    \includegraphics[trim = 2mm 5mm 2mm 5mm, clip, width=\linewidth]{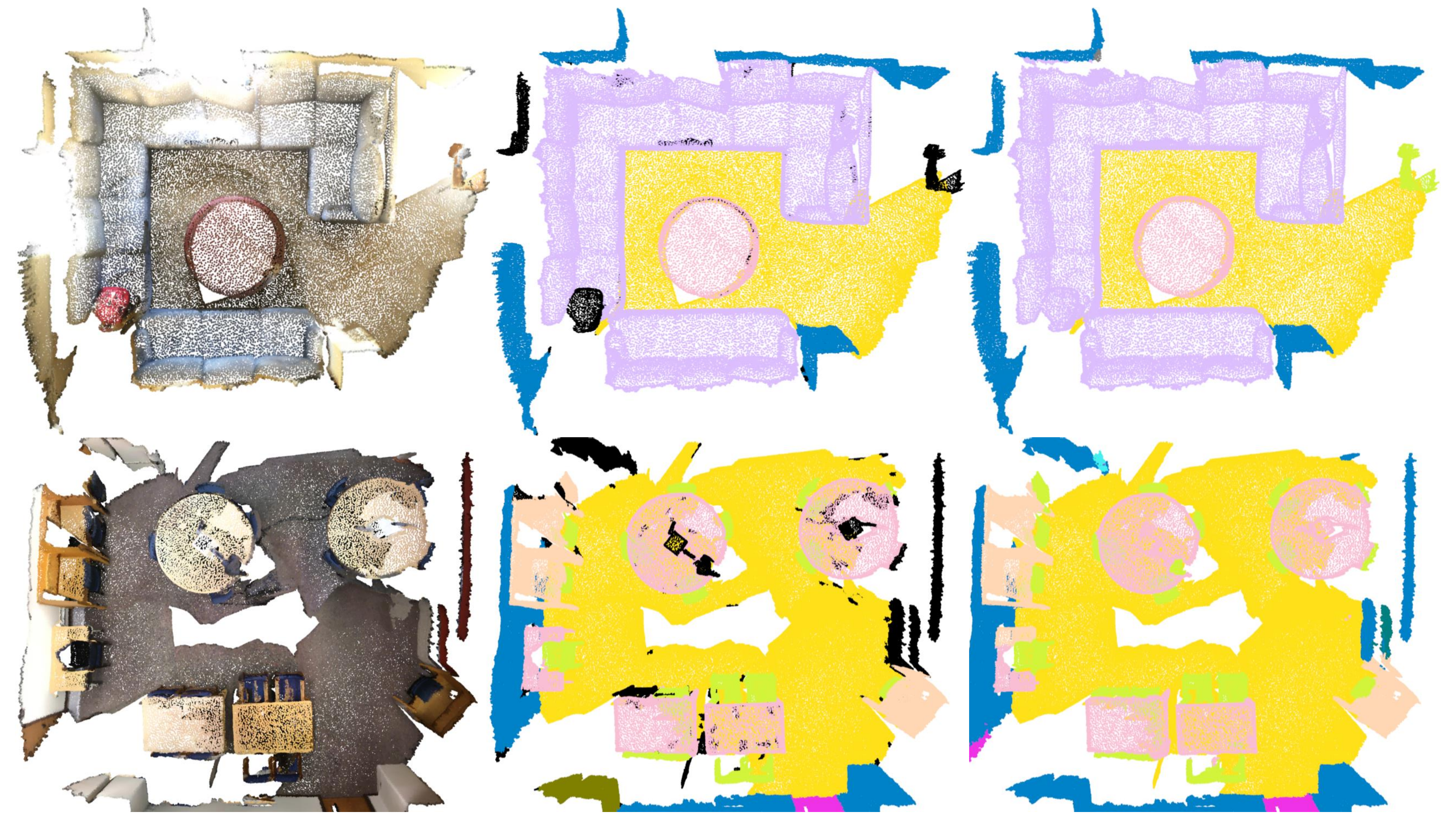}
    \includegraphics[trim = 10mm 90mm 10mm 90mm, clip, width=\linewidth]{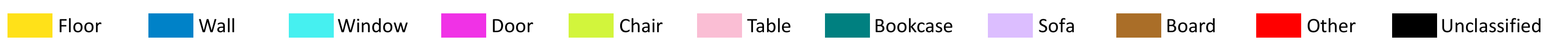}
    \caption{\textbf{Qualitative Results on ScanNet.} Sample qualitative results of SALANet with $S=2$ groups on ScanNet validation-set.}
    \label{fig:supqualscannet}
    \vspace{-2mm}
\end{figure*}

Table \ref{table:supscannet} shows the per class IoU on ScanNet test-set. SALANet with $S=2$ groups achieves competitive performance while using much less parameters in comparison with other SOTA methods with much more complex kernels. Sample qualitative results on validation-set are showcased in Figure \ref{fig:supqualscannet}.

\subsection{PartNet}
\begin{figure*}[htb]
    \centering
    \includegraphics[trim = 2mm 5mm 2mm 5mm, clip, width=\linewidth]{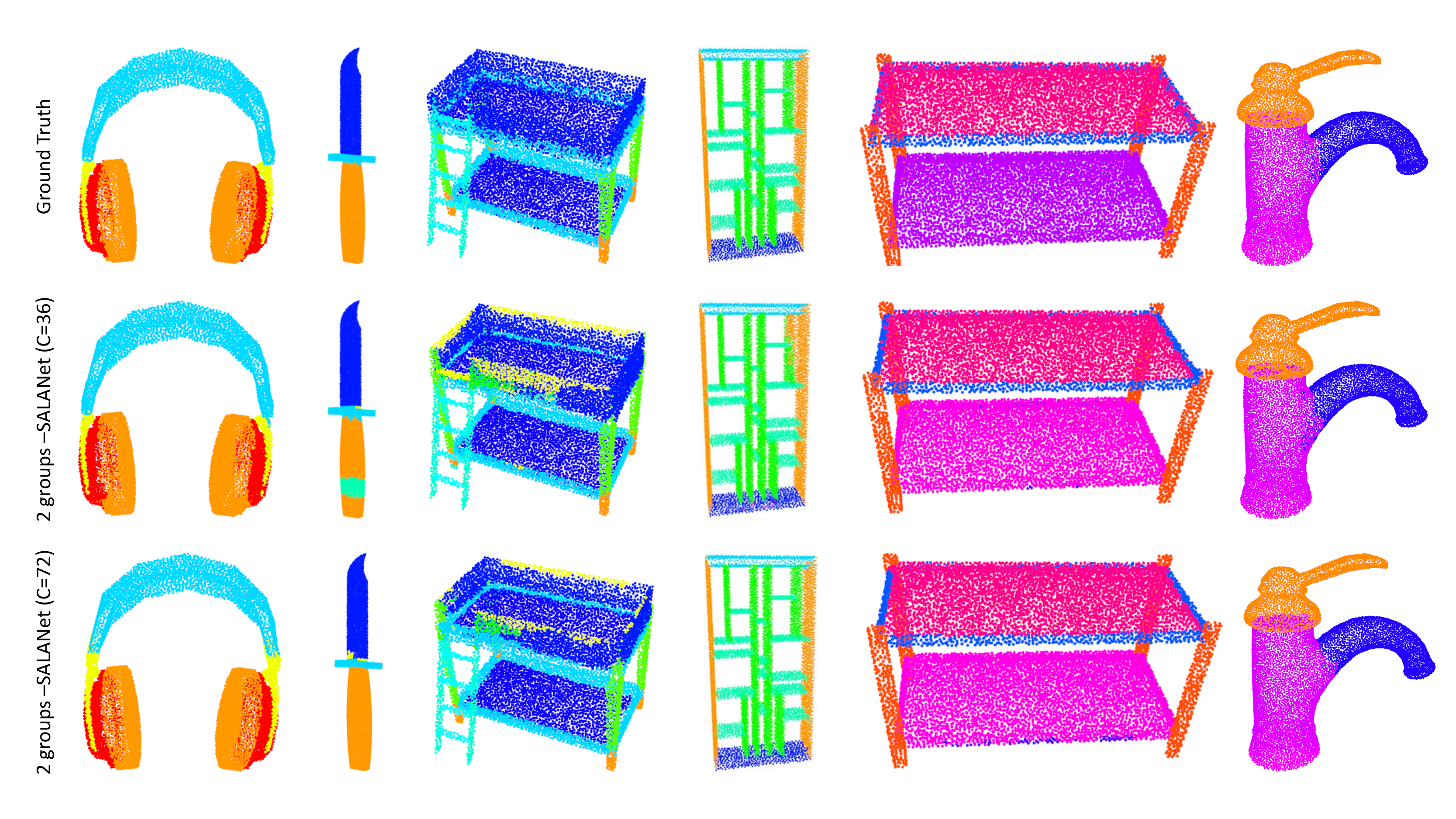}
    \caption{\textbf{Qualitative Results on PartNet.} Sample qualitative results of SALANet with $S=2$ groups on PartNet test-set using output channel dimension $C=36$ and $C=72$}
    \label{fig:supqualpartnet}
    \vspace{-2mm}
\end{figure*}

Table \ref{table:suppartnet} shows the per shape category mean part IoU (mpIoU) on PartNet test-set. \cite{closerlook3d} noted that increasing the output channel dimension $C$ is especially important on complex datasets such as PartNet. We vary the output channel dimension $C$ to test our model at different model learning capacities. Increasing the channel dimension compromises the parameter efficiency claim of SALANet. SALANet with $S=2$ groups achieves competitive performance in comparison with much larger models. Sample qualitative results on test-set are showcased in Figure \ref{fig:supqualpartnet}.


\begin{table*}[ht]
    \small
    \tabcolsep=0.08cm
    \begin{center}
        \resizebox{\textwidth}{!}{
        \begin{tabular}{l  | c | c c c c c c c c c c c c c}
            \hline
            \multicolumn{15}{c}{Semantic Segmentation on S3DIS Area 5}\\
            \hline
            Method & mIoU & ceiling & floor & wall & beam & column & window & door & table & chair & sofa & book. & board & clutter \\
            \hline
            PointNet \cite{pointnet} & 41.1 & 88.8 & 97.3 & 69.8 &  0 &  3.9  & 46.3 & 10.8 & 52.6 & 58.9 & 40.3  & 5.9  & 26.4 & 33.2 \\
            ResGCN28 \cite{deepgcns} & 52.5 & -     & -     & -     & -    & -     & -     & -     & -     & -     & -     & -     & -     & -     \\
            PointCNN \cite{pointcnn} & 57.3 & 92.3 & 98.2 & 79.4 & 0  & 17.6 & 22.8 & 62.1 & 74.4 & 80.6 & 31.7 &  66.7  & 62.1 & 56.7 \\
            SSP+SPG \cite{sspspg}                & 61.7  & -     & -     & -     & -    & -     & -     & -     & -     & -     & -     & -     & -     & -     \\
            SPH3D-GCN \cite{sph3d}              & 59.5  & 93.3  & 97.1  & 81.1  & 0    & 33.2  & 45.8  & 43.8  & 79.7  & 86.9  & 33.2  & 71.5  & 54.1  & 53.7  \\
            SegGCN \cite{seggcn}               & 63.6  & 93.7  & \textbf{98.6}  & 80.6  & 0    & 28.5  & 42.6  & \textbf{74.5}  & 80.9  & 88.7  & 69.0    & 71.3  & 44.4  & 54.3  \\
            PointASNL \cite{pointasnl}             & 62.6  & 94.3  & 98.4  & 79.1  & 0    & \textbf{26.7}  & 55.2  & 66.2  & 83.3  & 86.8  & 47.6  & 68.3  & 56.4  & 52.1  \\
            GACNet \cite{gacnet}                & 62.9  & 92.3  & 98.3  & 81.9  & 0    & 20.4  & \textbf{59.1}  & 40.9  & 78.5  & 85.8  & 61.7  & 70.8  & \textbf{74.7}  & 52.8  \\
            PointwiseMLP(25.6M) \cite{closerlook3d}    & 66.2  & 93    & 98.4 & 82.1  & 0 & 18.5 & \textbf{59.1}  & 63 & \textbf{83.5} & 91.1 & \textbf{76.4} & 76 & 62.5 & 57.3 \\
            Pseudo Grid (18.5M) \cite{closerlook3d}       & 65.9  & 94 & 98.4 & \textbf{82.6} & 0 & 20 & 55.7 & 65.2 & 81.5 & 91.5 & 65.3  & 76.3 & 66.7 & 60.1 \\
            PosPool*   (18.5M) \cite{closerlook3d}     & 66.7 & \textbf{94.9} & 98.4 & 82.5 & 0 & 24 & 51.8 & 70.6 & 82.3 & \textbf{92}    & 68.1 & 76.7 & 65 & 61.1 \\
            
            KPConv (r) (14.1M)\cite{kpconv}             & 65.4  & 92.6  & 97.3  & 81.4  & 0    & 16.5  & 54.5  & 69.5  & 80.2 & 90.1   & 66.4  & 74.6  & 63.7  & 58.1  \\
            KPConv (d) (14.9M)\cite{kpconv}             & 67.1  & 92.8  & 97.3  & 82.4  & 0    & 23.9  & 58    & 69 & 81.5    & 91      & 75.4  & 75.3  & 66.7  & 58.9  \\
            \hline 
            \hline
            2 groups - SALANet (1.6M)   & \textbf{67.6} & \textbf{94.9} & 98.3 & 82.5 & 0 & 26.2 & 55 & 71.1 & 82.7 & 91.3 & 73.3 & \textbf{77.8} & 63.8 & \textbf{61.9} \\
            \hline
        \end{tabular}}
        \vspace{3pt}
        \caption{\textbf{Scene semantic segmentation mIoU comparison on S3DIS Area 5}. SALANet with $S=2$ groups achieves SOTA on S3DIS Area $5$ while using close to $10\times$ less parameters than the reigning method. SALANet is trained and tested under the same settings in \cite{closerlook3d,kpconv}. $\mathbf{M}$ corresponds to million parameters.}\label{table:sups3disa5}
    \end{center}
\end{table*}

\begin{table*}[t]
    \small
    \tabcolsep=0.05cm
    \begin{center}
        \resizebox{\textwidth}{!}{
        \begin{tabular}{l | c | c | c| c c c c c c c c c c c c c}
            \hline
            \multicolumn{17}{c}{Semantic Segmentation on S3DIS 6-Fold-Cross-Validation}\\
            \hline
            Method & OA & mAcc & mIoU & ceiling & floor & wall & beam & column & window & door & table & chair & sofa & book. & board & clutter \\
            \hline
            PointCNN \cite{pointcnn} & 88.1 & 75.6 & 65.4 & 94.8 & 97.3 & 75.8 & 63.3 & 51.7 & 58.4 & 57.2 & 71.6 & 69.1 & 39.1 & 61.2 & 52.2 & 58.6\\
            SSP+SPG \cite{sspspg}            & 87.9   & 78.3   & 68.4  & -     & -     & -     & -     & -     & -     & -     & -     & -     & -     & -     & -     & -     \\
            SPH3D-GCN \cite{sph3d}           & 88.6   & 77.9   & 68.9  & 93.3  & 96.2  & 81.9  & 58.6  & 55.9  & 55.9  & 71.7  & 72.1  & 82.4  & 48.5  & 64.5  & 54.8  & 60.4  \\
            SegGCN \cite{seggcn}             & 87.8   & 77.1   & 68.5  & 92.5  & 97.6  & 78.9  & 44.6  & \textbf{58.2}  & 53.7  & 67.3  & 74.6  & 83.9  & 68    & 65.7  & 46.8  & 58.5  \\
            PointASNL \cite{pointasnl}          & 88.8   & 79     & 68.7  & \textbf{95.3}  & \textbf{97.9}  & 81.9  & 47    & 48    & \textbf{67.3}  & 70.5  & 71.3  & 77.8  & 50.7  & 60.4  & 63    & 62.8  \\
            RandLA-Net \cite{randlanet} & 88 & \textbf{82} & 70 & 93.1 & 96.1 & 80.6 & 62.4 & 48 & 64.4 & 69.4 & 69.4 & \textbf{76.4} & 60 & 64.2 & \textbf{65.9} & 60.1 \\
            KPConv (r)(14.1M) \cite{kpconv}  & -      & 78.1   & 69.6  & 93.7  & 92    & 82.5  & 62.5  & 49.5  & 65.7  & 77.3  & 64    & 57.8  & 71.7  & 68.8  & 60.1  & 59.6  \\
            KPConv (d)(14.9M) \cite{kpconv}   & -      & 79.1   & 70.6  & 93.6  & 92.4  & \textbf{83.1}  & 63.9  & 54.3  & 66.1  & 76.6  & 57.8  & 64    & \textbf{69.3}  & \textbf{74.9}  & 61.3  & 60.1  \\
            \hline
            \hline
            
            2 groups - SALANet (1.6M) & \textbf{89.9}  & 80.6  & \textbf{72.5} & 94.6 & 96.6 & 82.6 & \textbf{66.9} & 51.7 & 66.3 & \textbf{79.2} & \textbf{75.6} & 75.7 & 62.9 & 71.3 & 54.9  & \textbf{64.5} \\
            \hline
            
        \end{tabular}}
        \vspace{1pt}
        \caption{\textbf{Scene semantic segmentation mIoU comparison on S3DIS 6-fold-cross-validation}. SALANet with $S=2$ groups achieves SOTA on S3DIS 6-fold-cross-validation benchmark, outperforming the reigning method with almost 2 mIoU points while using close to $10\times$ less parameters. SALANet is trained and tested under the same settings in \cite{closerlook3d,kpconv}. $\mathbf{M}$ corresponds to million parameters.}\label{table:sups3dis6fold}
    \end{center}
\end{table*}



\begin{table*}[t]
    \small
    \tabcolsep=0.06cm
    \begin{center}
        \resizebox{\textwidth}{!}{
        \begin{tabular}{l|c|cccccccccccccccccccc}
        \hline
        \multicolumn{22}{c}{Semantic Segmentation on ScanNet} \\
        \hline
        Methods & mIoU & floor & wall & chair & sofa & table & door & cab & bed & desk & toil & sink & wind & pic & bkshlf & curt & show & cntr & fridg & bath & other \\
        \hline
        PointCNN \cite{pointcnn} & 45.8 & 94.4 &  70.9 &  71.5 &  54.5 &  45.6 &  31.9 &  32.1 &  61.1 &  32.8 &  75.5 &  48.4 &  47.5 &  16.4 &  35.6 &  37.6 &  22.9 &  29.9 &  21.6 &  57.7 &  28.5 \\
        SPH3D-GCN \cite{sph3d} & 61 & 93.5 & 77.3 & 79.2 & 70.5 & 54.9 & 50.7 & 53.2 & 77.2 & 57 & 85.9 & 60.2 & 53.4 & 4.6 & 48.9 & 64.3 & 70.2 & 40.4 & 51 & \textbf{85.8} & 41.4 \\
        SegGCN\cite{seggcn} & 58.9 & 93.6 & 77.1 & 78.9 & 70 & 56.3 & 48.4 & 51.4 & 73.1 & 57.3 & 87.4 & 59.4 & 49.3 & 6.1 & 53.9 & 46.7 & 50.7 & 44.8 & 50.1 & 83.3 & 39.6 \\

        PointASNL \cite{pointasnl} & 66.6 & \textbf{95.1} & 80.6 & \textbf{83.0} & 75.1 & 55.3 & 53.7 & \textbf{65.5} & \textbf{78.1} & 47.4 & 81.6 & 67.5 & \textbf{70.3} & \textbf{27.9} & 75.1 & 76.9 & 69.8 & 47.1 & \textbf{63.5} & 70.3 & \textbf{47.5} \\
        KPConv (14.9M) \cite{kpconv} & \textbf{68.4} & 93.5 & \textbf{81.9} & 81.4 & \textbf{78.5} & 61.4 & \textbf{59.4} & 64.7 & 75.8 & 60.5 & 88.2 & \textbf{69} & 63.2 & 18.1 & \textbf{78.4} & \textbf{77.2} & 80.5 & \textbf{47.3} & 58.7 & 84.7 & 45 \\
        \hline
        \hline
        2 groups - SALANet (1.6M) & 67.0 & 92.4 & 79.4 & 80.7 & 74.6 & \textbf{62.3} & 54.5 & 65.2 & 77.0 & \textbf{65.9} & \textbf{89.2} & 63.5 & 57.0 & 14.9 & 76.8 & 74.7 & \textbf{81.1} & 45.1 & 57.1 & 81.6 & 47.3\\
        \hline
        \end{tabular}}%
        \vspace{1pt}
        \caption{\textbf{Scene semantic segmentation mIoU comparison on ScanNet}. SALANet with $S=2$ groups achieves competitive performance with much bigger and complex kernels on ScanNet test set while using almost $10\times$ less parameters. SALANet is trained and tested under the same settings in \cite{kpconv}. $\mathbf{M}$ corresponds to million parameters.}\label{table:supscannet}
    \end{center}
\end{table*}



\begin{table*}[t]
    \small
    \tabcolsep=0.03cm
    \begin{center}
        \resizebox{\textwidth}{!}{
        \begin{tabular}{l | c | c  c c c c c c c c c c c c c c c c}
            \hline
            \multicolumn{19}{c}{Part Segmentation on PartNet}\\
            \hline
            Methods & mpIoU & Bed & Bottle & Chair & Clock & Dishws & Display & Door & Earph & Faucet & Knife & Lamp & Microw & Refriger & StrgFur & Table & TrshCan & Vase \\
            \hline 
            PointNet \cite{pointnet} & 35.6 & 13.4 & 29.5 & 27.8 & 28.4 & 48.9 & 76.5 & 30.4 & 33.4 & 47.6 & 32.9 & 18.9 & 37.2 & 33.5 & 38 & 29 & 34.8 & 44.4 \\
            PointNet++ \cite{pointnet++} & 42.5 & 30.3 & 41.4 & 39.2 & 41.6 & 50.1 & 80.7 & 32.6 & 38.4 & 52.4 & 34.1 & 25.3 & 48.5 & 36.4 & 40.5 & 33.9 & 46.7 & 49.8 \\

            ResGCN28 \cite{deepgcnsext} & 45.1 & 35.9 & 49.3 & 41.1 & 33.8 & 56.2 & 81 & 31.1 & 45.8 & 52.8 & 44.5 & 23.1 & 51.8 & 34.9 & 47.2 & 33.6 & 50.8 & 54.2 \\
            PointCNN \cite{pointcnn} & 46.5 & 41.9 & 41.8 & 43.9 & 36.3 & 58.7 & 82.5 & 37.8 & 48.9 & 60.5 & 34.1 & 20.1 & 58.2 & 42.9 & 49.4 & 21.3 & 53.1 & 58.9\\
            PointwiseMLP(C=36, 1.6M) \cite{closerlook3d} & 47.0 & - & - & - & - & - & - & - & - & - & - & - & - & - & - & - & - & - \\

            Pseudo   Grid (C=36, 1.2M) \cite{closerlook3d} & 45.2 & - & - & - & - & - & - & - & - & - & - & - & - & - & - & - & - & - \\
            PosPool*   (C=36, 1.1M) \cite{closerlook3d} & 47.2 & - & - & - & - & - & - & - & - & - & - & - & - & - & - & - & - & - \\
            PointwiseMLP(C=144, 25.6M) \cite{closerlook3d} & 51.2 & 44.5 & 52.6 & 46.0 & 38.4 & \textbf{68.2} & 82.5 & 46.9 & 47.1 & 58.7 & 43.8 & 26.4 & 59.2 & 48.7 & 52.5 & 41.3 & 55.4 & 57.3 \\
            Pseudo Grid (C=144, 18.5M) \cite{closerlook3d}& 53 & 47.5 & 50.9 & \textbf{49.2} & 44.9 & 67 & \textbf{84.2} & 49.1 & 49.9 & \textbf{62.8} & 38.3 & \textbf{27} & 59.4 & 54.3 & 54.1 & 44.5 & 57.4 & 60.7 \\
            PosPool*   (C=144, 18.5M)\cite{closerlook3d} & \textbf{53.8} & \textbf{49.5} & 49.4 & 48.3 & \textbf{49} & 65.6 & \textbf{84.2} & \textbf{56.8} & \textbf{53.8} & 62.4 & 39.3 & 24.7 & \textbf{61.3} & \textbf{55.5} & \textbf{54.6} & \textbf{44.8} & 56.9 & 58.2 \\
            \hline
            \hline
            2 groups - SALANet (C=36, 1.6M) & 49.4 & 46.5 & 44.7 & 45 & 44.7 & 65.0 & 81.7 & 39.9 & 49.5 & 50.6 & 44.2 & 26.3 & 54.6 & 49.0 & 47.8 & 39.5 & 51.4 & 59.4 \\
            2 groups - SALANet (C=72, 6.4M) & 52.4 & 46.7 & 51.6 & 46.2 & 42.5 & 66.8 & 84.0 & 46.9 & 53.1 & 57.8 & \textbf{46.0} & 26.4 & 57.3 & 52.9 & 52.3 & 42.7 & 56.3 & \textbf{61.3} \\
            2 groups - SALANet (C=90, 10M) & 53.1 & 46.9 & \textbf{55.7} & 45.8 & 48.5 & 67.4 & 82.8 & 50.5 & 50.1 & 57.6 & 45.8 & 25.8 & 58.4 & 54.8 & 53.0 & 43.3 & \textbf{57.4} & 59.5\\
            \hline
        \end{tabular}}
        \vspace{3pt}
        \caption{\textbf{Shape part segmentation mpIoU comparison on PartNet}. SALANet with $S=2$ groups and $C=72$ achieves competitive performance on PartNet (level 3) test set in comparison with much bigger models. SALANet is trained and tested under the same settings in \cite{closerlook3d}. $\mathbf{M}$ corresponds to million parameters.}\label{table:suppartnet}
    \end{center}
\end{table*}



\clearpage

\clearpage

\clearpage

\end{document}